\documentclass[sigconf]{acmart}
\usepackage{dirtytalk}

\AtBeginDocument{%
  \providecommand\BibTeX{{%
    \normalfont B\kern-0.5em{\scshape i\kern-0.25em b}\kern-0.8em\TeX}}}

\copyrightyear{2023}
\acmYear{2023}
\setcopyright{rightsretained}
\acmConference[AIES '23]{AAAI/ACM Conference on AI, Ethics, and Society}{August 8--10, 2023}{Montréal, QC, Canada}
\acmBooktitle{AAAI/ACM Conference on AI, Ethics, and Society (AIES '23), August 8--10, 2023, Montréal, QC, Canada}
\acmDOI{10.1145/3600211.3604754}
\acmISBN{979-8-4007-0231-0/23/08}
\begin{document}

\title{Towards a Holistic Approach: Understanding Sociodemographic Biases in NLP Models using an Interdisciplinary Lens}


\author{Pranav Narayanan Venkit}
\affiliation{
  \institution{Pennsylvania State University}
  \city{University Park}
  \state{Pennsylvania}
  \country{USA}
  \postcode{16803}
}
\email{pranav.venkit@psu.edu}
\orcid{0000-0002-5671-0461}

\renewcommand{\shortauthors}{Pranav Narayanan Venkit}

\begin{abstract}

The rapid growth in the usage and applications of Natural Language Processing (NLP) in various sociotechnical solutions has highlighted the need for a comprehensive understanding of bias and its impact on society. While research on bias in NLP has expanded, several challenges persist that require attention. These include the limited focus on sociodemographic biases beyond race and gender, the narrow scope of analysis predominantly centered on models, and the technocentric implementation approaches.

This paper addresses these challenges and advocates for a more interdisciplinary approach to understanding bias in NLP. The work is structured into three facets, each exploring a specific aspect of bias in NLP. The \textbf{first facet} focuses on identifying sociodemographic bias in various NLP architectures, emphasizing the importance of considering both the models themselves and human computation to comprehensively understand and identify bias. In the \textbf{second facet}, we delve into the significance of establishing a shared vocabulary across different fields and disciplines involved in NLP. By highlighting the potential bias stemming from a lack of shared understanding, this facet emphasizes the need for interdisciplinary collaboration to bridge the gap and foster a more inclusive and accurate analysis of bias. Finally, the \textbf{third facet} investigates the development of a holistic solution by integrating frameworks from social science disciplines. This approach recognizes the complexity of bias in NLP and advocates for an interdisciplinary framework that goes beyond purely technical considerations, involving social and ethical perspectives to address bias effectively.

The \textbf{first facet} includes the following of my published works \cite{venkit2021identification, venkit2022study, venkit2023nationality, venkit2023unmasking} to provide results into how the importance of understanding the presence of bias in various minority group that has not been in focus in the prior works of bias in NLP. The work also shows the need to create a method that considers both human and AI indicators of bias, showcasing the importance of the first facet of my research. 
In my study \cite{venkit2021identification}, I delve into sentiment analysis and toxicity detection models to identify explicit bias against race, gender, and people with disabilities (PWDs). Through statistical exploration of conversations on social media platforms such as Twitter and Reddit, I gain insights into how disability bias permeates real-world social settings. To quantify explicit sociodemographic bias in sentiment analysis and toxicity analysis models, I create the Bias Identification Test in Sentiment (BITS) corpus\footnote{https://github.com/PranavNV/BITS}. Applying BITS, I uncover significant biases in popular AIaaS sentiment analysis tools, including TextBlob, VADER, and Google Cloud Natural Language API, as well as toxicity analysis models like Toxic-BERT. Remarkably, all of these models exhibit statistically significant explicit bias against disability, underscoring the need for comprehensive understanding and mitigation of biases affecting such groups. The work also demonstrates the utility of BITS as a model-independent method of identifying bias by focusing on social groups instead.

Expanding on this, my next work \cite{venkit2022study} delves into the realm of \textit{implicit bias} in NLP models. While some models may not overtly exhibit bias, they can unintentionally perpetuate harmful stereotypes \cite{dev2021onmeasures}. To measure and identify implicit bias in commonly used embedding and large language models, I propose a methodology to measure social biases in various NLP architectures. Focusing on people with disabilities (PWD) as a group with complex social dynamics, I analyze various word embedding-based and transformer-based LLMs, revealing significant biases against PWDs in all tested models. These findings expose how models trained on extensive corpora tend to favor ableist language, underscoring the urgency of detecting and addressing implicit bias. The above two works look at both the implicit and explicit nature of bias in NLP, showcasing the need to distinguish the efforts placed in understanding them. The results also demonstrate the utility of identifying such biases as it provides context to the black-box nature of such public models.

As the field of NLP evolved from embedding-based models to large language models, the way these models are constructed underwent significant changes \cite{radford2019language}. However, the concern arises from the fact that these models often reflect a \textit{populist} viewpoint \cite{bender2021dangers} that perpetuates majority-held ideas rather than objective truths. This difference in perception can lead to biases perpetuated by the majority's worldview. To explore this aspect, I investigate how LLMs represent nationality and their impact on societal stereotypes \cite{venkit2023nationality}. By examining LLM-generated stories for various nationalities, I establish a correlation between sentiment and the population of internet users in a country. The study reveals the unintentional implicit and explicit nationality biases exhibited by GPT-2, with nations having lower internet representation and economic status generating negative sentiment stories and employing a greater number of negative adjectives. Additionally, I explore potential debiasing methods such as adversarial triggering and prompt engineering, demonstrating their efficacy in mitigating stereotype propagation through LLM models. 

While prior work predominantly relies on automatic indicators like sentiment scores or vector distances to identify bias \cite{bolukbasi2016man}, the next phase of my research emphasizes the importance of understanding biases through the lens of human readers \cite{venkit2023unmasking}, bringing to light the need for a human lens in understanding bias through human-aided indicators and mixed-method identification. By incorporating concepts of social computation, using human evaluation, we gain a better understanding of biases' potential societal impact within the context of language models. To achieve this, I conduct open-ended interviews and employ qualitative coding and thematic analysis to comprehend the implications of biases on human readers. The findings demonstrate that biased NLP models tend to replicate and amplify existing societal biases, posing potential harm when utilized in sociotechnical settings. The qualitative analysis from the interviews provides valuable insights into readers' experiences when encountering biased articles, highlighting the capacity to shift a reader's perception of a country. These findings emphasize the critical role of public perception in shaping AI's impact on society and the need to correct biases in AI systems.

The \textbf{second facet} of my research aims to bridge the disparity between AI research and society. This disparity has resulted in a lack of shared understanding between these domains, leading to potential biases and harm toward specific groups. Employing an interdisciplinary approach that combines social informatics, philosophy, and AI, I will investigate the similarities and disparities in the concepts utilized by machine learning models. Existing research \cite{blodgett2020language} highlights the insufficient interdisciplinary effort and motivation in comprehending social aspects of NLP. To commence this exploration, I will delve into the shared taxonomy of \textit{sentiment} and \textit{fairness} in natural language processing, sociology, and humanities. This research will first delve into the interdisciplinary nature of sentiment and its application in sentiment analysis models. Sentiment analysis, a popular machine learning application for text classification based on sentiment, opinion, and subjectivity, holds significant influence as a sociotechnical system that impacts both social and technical actors within a network. Nevertheless, the definition and connotation of sentiment vary vastly across different research fields, potentially leading to misconceptions regarding the utility of such systems. To address this issue, this study will examine how diverse fields, including psychology, sociology, and technology, define the concept of sentiment. By unraveling the divergent perspectives on sentiment within different fields, the paper will uncover discrepancies and varying applications of this interdisciplinary concept. Additionally, the research will survey commonly utilized sentiment analysis models, aiming to comprehend their standardized definitions and associated issues. Ultimately, the study will pose critical questions that should be considered during the development of social models to mitigate potential biases and harm stemming from an insufficiently defined comprehension of fundamental social concepts. Similar efforts will be dedicated to comprehending the disparity in bias and fairness as an interdisciplinary concept, shedding light on the imperative for inclusive research to cultivate superior AI models as sociotechnical solutions.

The \textbf{third facet} of my study embarks upon an exploration of the intricate interplay between human and AI actors, employing the formidable theoretical lens of actor-network theory (ANT). Through the presentation of a robust framework, this facet aims to engender the formation of efficacious development networks that foster collaboration among developers, practitioners, and other essential stakeholders. Such inclusive networks serve as crucibles for the cultivation of holistic solutions that transcend the discriminatory trappings afflicting specific populations. A tangible outcome of this endeavor entails the creation of an all-encompassing bias analysis platform, poised to guide the discernment and amelioration of an array of sociodemographic biases manifesting within any machine-learning system. By catalyzing the development of socially aware and less pernicious technology, this research makes a substantial contribution to the realms of NLP and AI.

The significance of this proposed research reverberates beyond the confines of NLP, resonating throughout the broader domain of AI, wherein analogous challenges about social biases loom large. Leveraging the proposed framework, developers, practitioners, and policymakers are empowered to forge practical solutions that embody inclusivity and reliability, especially when used as a service (AIaaS). Moreover, the platform serves as a centralized locus for the identification and rectification of social biases, irrespective of the underlying model or architecture. By furnishing a cogent narrative that underscores the imperative for a comprehensive and interdisciplinary approach, my work strives to propel the ongoing endeavors to comprehend and mitigate biases within the realm of NLP. With its potential to augment the equity, inclusivity, and societal ramifications of NLP technologies, the proposed framework catapults the field towards responsible and ethical practices.

\end{abstract}





\maketitle

\bibliographystyle{ACM-Reference-Format}
\bibliography{sample-base}


\end{document}